\documentclass[twocolumn,showpacs,%
  nofootinbib,aps,superscriptaddress,%
  eqsecnum,prd,notitlepage,showkeys,10pt]{revtex4-1}

\usepackage{amsmath}
\usepackage{amsfonts}
\usepackage{amssymb}

\usepackage{lipsum}
\usepackage{url}
\makeatletter
\def\ps@pprintTitle{%
 \let\@oddhead\@empty
 \let\@evenhead\@empty
 \def\@oddfoot{\centerline{\thepage}}%
 \let\@evenfoot\@oddfoot}
\makeatother

\usepackage{silence}
\usepackage{graphicx}
\usepackage[outdir=./]{epstopdf}
\usepackage{amssymb}
\usepackage{amsmath}
\usepackage{bm}
\usepackage{booktabs}
\usepackage{float}
\usepackage{enumitem}

\usepackage[caption=false]{subfig}

\usepackage{bookmark}

\usepackage{silence}
\def\code#1{\texttt{#1}}

\newcommand{\PLH}{{\mkern-2mu\times\mkern-2mu}}

\usepackage[margin=0.6in]{geometry}
\usepackage{titlesec}
\titlespacing*{\subsection}{0pt}{1.1\baselineskip}{\baselineskip}

\usepackage{titlesec}
\titlespacing*{\section}{0pt}{1.5\baselineskip}{\baselineskip}

\begin{document}

\title{Exploring the Deep Feature Space of a Cell Classification Neural Network}

\author{Ezra Webb}
\affiliation{\footnotesize Department of Chemistry, University of Tokyo, Tokyo, Japan}
\affiliation{\footnotesize Department of Physics, University of Cambridge, Cambridge, UK}
\author{Cheng Lei}
\affiliation{\footnotesize Department of Chemistry, University of Tokyo, Tokyo, Japan}
\author{Chun-Jung Huang}
\affiliation{\footnotesize Department of Chemistry, University of Tokyo, Tokyo, Japan}
\affiliation{\footnotesize Department of Photonics, National Chiao Tung University, Hsinchu, Taiwan}
\author{Hirofumi Kobayashi}
\affiliation{\footnotesize Department of Chemistry, University of Tokyo, Tokyo, Japan}
\author{Hideharu Mikami}
\affiliation{\footnotesize Department of Chemistry, University of Tokyo, Tokyo, Japan}
\author{Keisuke Goda}
\affiliation{\footnotesize Department of Chemistry, University of Tokyo, Tokyo, Japan}
\affiliation{\footnotesize Japan Science and Technology Agency, Tokyo, Japan}

\begin{abstract}
\noindent In this paper, we present contemporary techniques for visualising the feature space of a deep learning image classification neural network. These techniques are viewed in the context of a feed-forward network trained to classify low resolution fluorescence images of white blood cells captured using optofluidic imaging. The model has two output classes corresponding to two different cell types, which are often difficult to distinguish by eye. This paper has two major sections. The first looks to develop the information space presented by dimension reduction techniques, such as t-SNE, used to embed high-dimensional pre-softmax layer activations into a two-dimensional plane. The second section looks at feature visualisation by optimisation to generate feature images representing the learned features of the network. Using and developing these techniques we visualise class separation and structures within the dataset at various depths using clustering algorithms and feature images; track the development of feature complexity as we ascend the network; and begin to extract the features the network has learnt by modulating single-channel feature images with up-scaled neuron activation maps to distinguish their most salient parts.
\end{abstract}
\maketitle

\section{Introduction}

Deep learning methods can quite easily be viewed as black boxes, with our focus only being on hyperparameter tuning and the classification accuracy we ultimately achieve. Often, and perhaps for good reason, understanding exactly what a network is learning, and where its outputs come from, is not our primary goal. Here however it is, where in the context of a convolutional neural network used for cell classification, we look to understand what features our network has learnt, and how class separation occurs.

In this paper, we present a variety of techniques for exploring the feature space of a deep learning image classification network. The network we analyse uses low resolution fluorescence images, captured using a frequency-division-multiplexed fluorescence imaging flow cytometer \cite{Nitta, Mikami}, to classify two types of white blood cell. Over the past few years visualisation techniques have seen rapid development \cite{UMAP, lucid, feature-vis, deepvis, invert, inceptionism, deepdreaming, synthgen, building-blocks}, allowing us to probe more deeply than ever before into the workings of a convolutional neural network.

Section \ref{S:1} of this paper describes the model used in our analysis, our image processing techniques, and some basic approaches to visualising our network's feature space. Section \ref{S:2} explores dimension reduction techniques and how we can enhance the information space they present to better understand class separation and our dataset's structure. Section \ref{S:3} looks at feature visualisation by optimisation to directly visualise and interpret the learned features of the network. Section \ref{S:4} then briefly outlines attribution techniques. Finally, in section \ref{S:5} we present our conclusions.

\section{Network \& Image Data}
\label{S:1}

\subsection{Model Architecture} \label{ssec:Model Architecture}

\begin{figure}[h!]
    \centering
    \subfloat[]{{\includegraphics[width=4.6cm,trim=0.2cm 0cm -0.7cm 0.49cm]{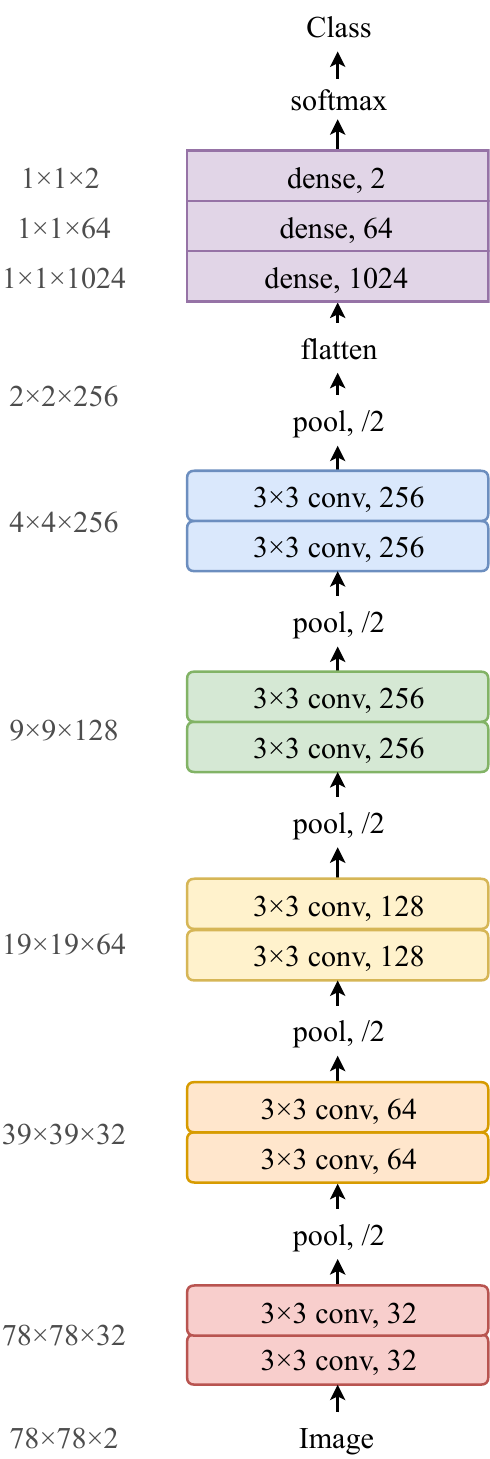} }}%
    \qquad
    \subfloat[]{{\includegraphics[width=2.63cm,trim=3.2cm 8.7cm 2cm 11.5cm]{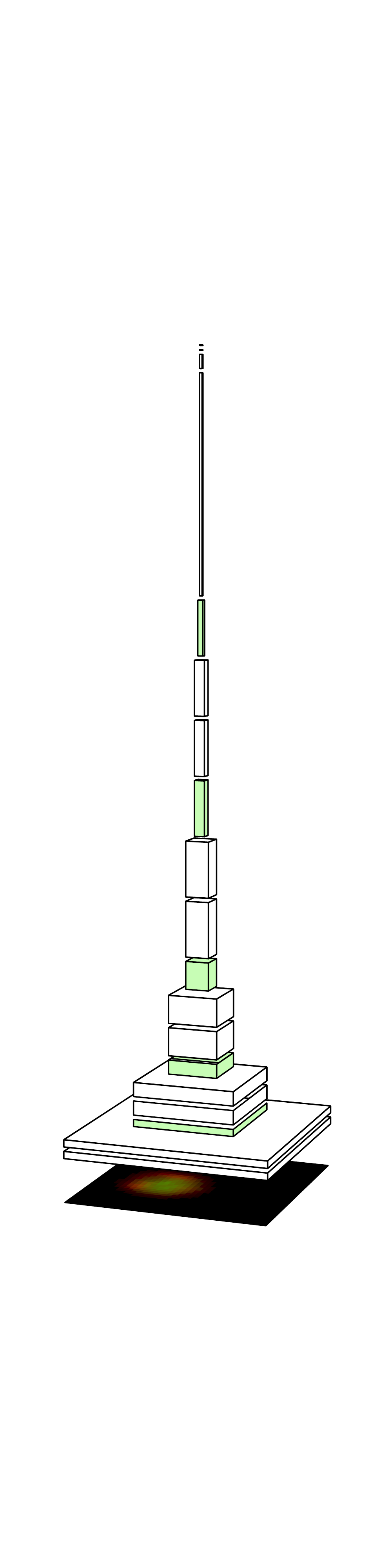} }}%
    \caption{\footnotesize Model Structure. (a) Schematic of the model architecture, where the dimensions in grey after each pooling layer indicate the shape the feature tensor. (b) Schematic showing how the shape of feature tensor changes as it moves up the network. The shape after a max-pooling layer is shown in green. Note the vertical axis is squashed relative the other two axes.}%
    \label{fig:architecture}%
\end{figure}

The model used in this investigation has a deliberately simple feed-forward architecture; consisting only of convolutional layers, max pooling layers and densely-connected layers. The full architecture is shown in figure \ref{fig:architecture}a and is effectively a scaled down version of the architecture of the VGG16 model \cite{VGG16}. It has a total of 3,467,746 trainable parameters and when saved as a protocol buffer file (including all parameters) is under 14MB.

\subsection{Cell Images} \label{ssec:Cell Images}
This analysis looks at fluorescence images of two different types of white blood cells: lymphocyte cells and neutrophil cells. The images of these cells were captured using a frequency-division-multiplexed fluorescence imaging flow cytometer \cite{Nitta, Mikami}, where for each cell two images were taken, corresponding to the two different florescent markers used. These two images were then combined into a single false-colour image, where broadly speaking the red channel shows the cytoplasm and the green channel shows the nucleus.

\begin{figure}
\includegraphics[width=9cm ,trim=3.7cm 1cm 2.8cm 0.1cm]{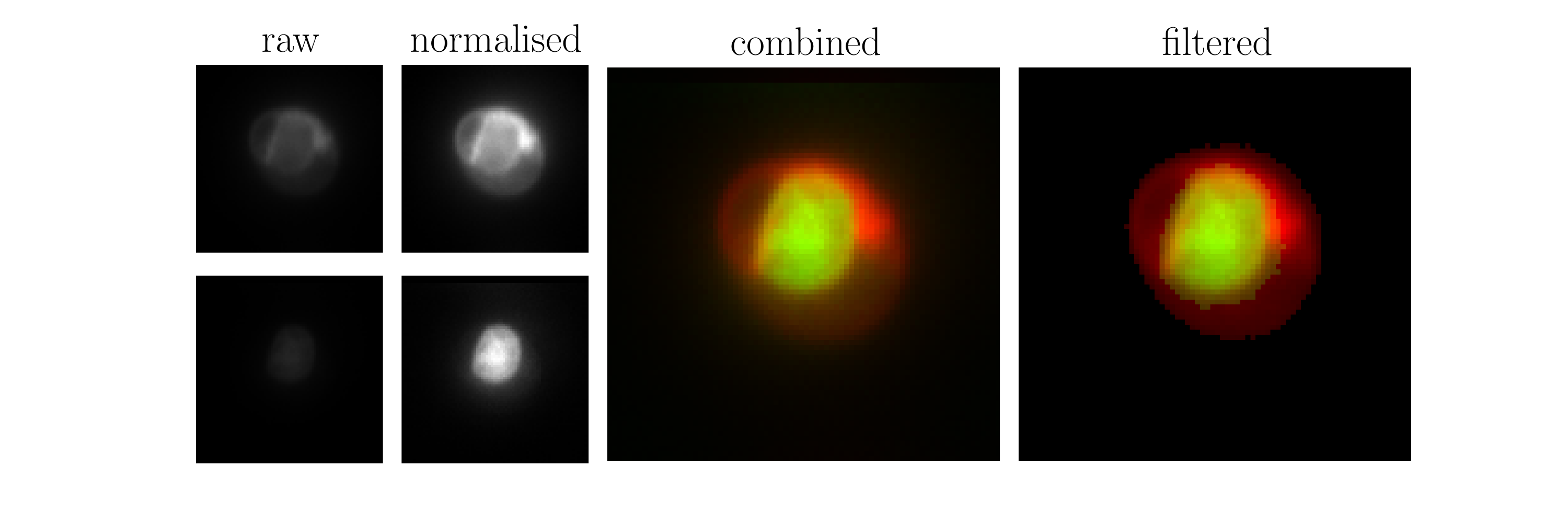}
\caption{\footnotesize Channel normalisation, combination and filtering. Left: images showing the two fluorescence channels of a lymphocyte cell before and after normalisation. Right: images of the filtered and unfiltered false-colour images of this cell.}
\label{fig:image_filter}%
\end{figure}

To prevent information leak of the cell type into the image due to the experimental set-up, the background of each cell image was removed. The exact approach to doing this is subjective and can have a sizeable effect on the classification accuracy. Here a quite strict approach to filtering the cell images was used, as achieving a high classification accuracy was not the main goal of this investigation. For our purposes, it is preferable to remove some of the actual cell information rather than run the risk of preserving some information about the experimental conditions. If some of this information was preserved, the neural network will potentially use this as a way to classify the images, and it may then manifest itself in the feature images generated in section \ref{S:3}, distorting our results.

The background of each fluorescence channel was removed by deleting any pixels with values less than 20\% of the channel maximum. After then combining the two channels into a single false-colour image, a number of additional conditions were checked to ensure these images were high quality. If an image passed these checks, each channel was then normalised and the image was added to the filtered image-set. This process is shown in figure \ref{fig:image_filter}. The final filtered image-set contained 10,901 lymphocyte cell images and 17,424 neutrophil cell images. These images were then split into training, validation and test datasets in a 60:20:20 ratio.

\subsection{Training \& Accuracy}\label{ssec:training}

In training the network, only horizontal flips and 90$^{\circ}$ rotations were used to transform the images, as transformations which distorted the pixel values had a significant impact on classification accuracy. Optimisation was done using stochastic gradient decent with momentum, and a 25\% dropout was used after each max-pooling pooling layer.

\begin{figure}
\includegraphics[width=7cm ,trim=0.2cm -0.2cm 0cm 1cm]{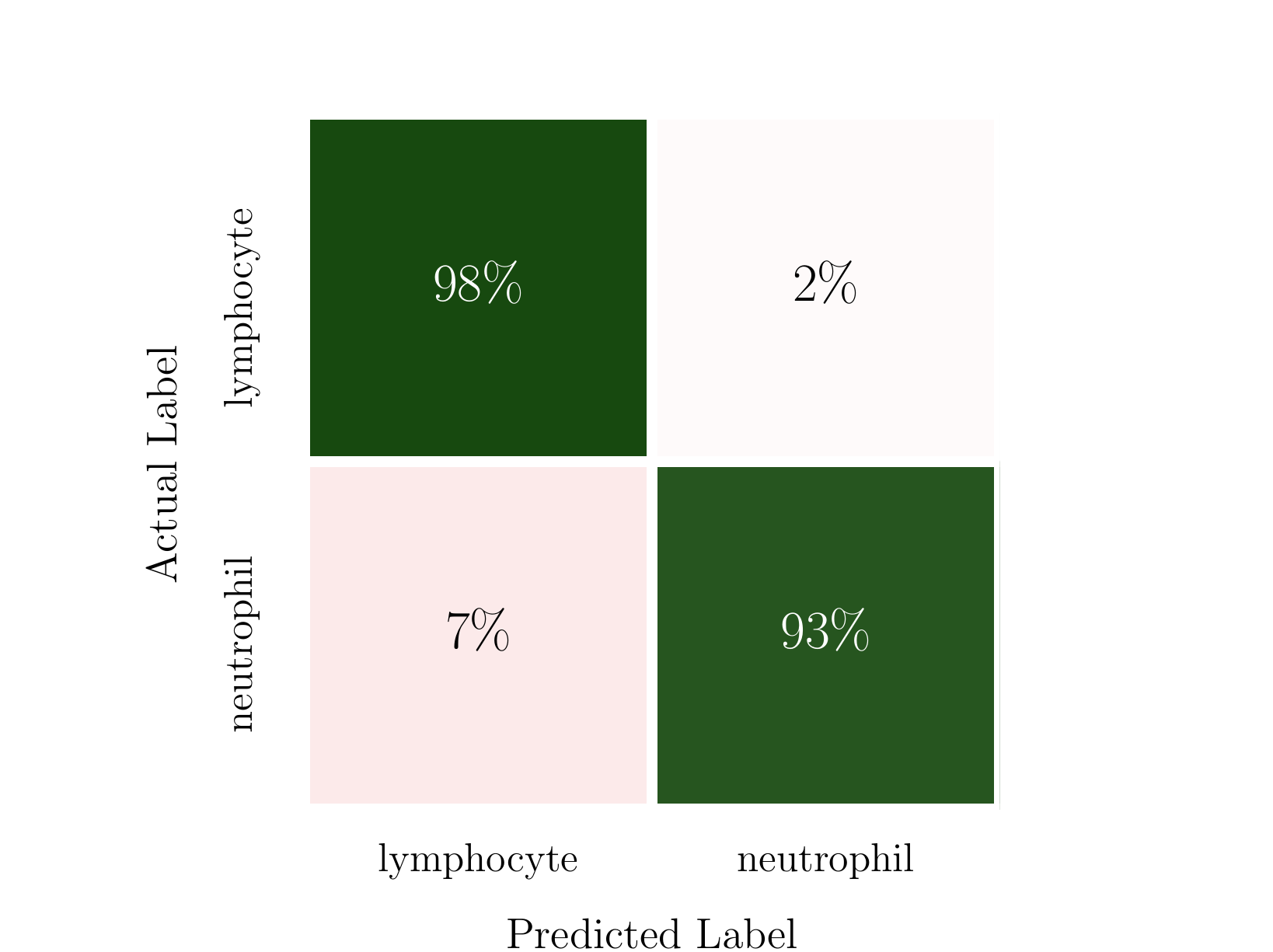}
\caption{\footnotesize Classification accuracy of our network on the test image-set (containing 2180 lymphocyte cell images and 3445 neutrophil cell images). Overall accuracy was 95\%.}
\label{fig:hinton}%
\end{figure}

As extremely high classification accuracy was not the goal of this investigation, the accuracies shown in figure \ref{fig:hinton} were deemed reasonable. Higher classification accuracies ($\sim$ 97 - 98\%) were achieved when less aggressive filtering techniques were used.

\subsection{The Feature Tensor} \label{ssec:language}

The feature tensor is a three-dimensional grid of values which represent a cell image as it moves through the network. The neural network operates on and transforms this grid of values as it moves up the layers. At a given layer, each value in the feature tensor represents the output of a neuron connected to the previous layer. Before entering the neural network, a cell image (here a 78$\PLH$78 pixel two channel image) is a tensor with dimensions 78$\PLH$78$\PLH$2, corresponding to height, width and depth. As this grid of values ascends the convolutional section of our network, its height and width are reduced, and its depth is increased. In our model, the tensor's depth increases to a maximum of 256 units, and as such becomes abstracted from our conventional notion of an `image'. However, it can be still useful to think of the feature tensor as an object similar to an image; understanding the different depth-wise unit slices to be different channels of some `complex image', with each representing some different aspect of the cell over a 2D space. Therefore, we will henceforth refer to these depth-wise unit slices of the feature tensor as channels, and refer to the remaining two dimensions as the spatial dimensions of the feature tensor.\footnote{Note that what we refer to here as \textit{channels} are sometimes alternatively referred to as \textit{filters} or \textit{features} in the literature.} Figure \ref{fig:architecture}b is a schematic showing how the dimensions of the feature tensor change as it passes through the layers of our network, starting as a 78$\PLH$78$\PLH$2 image, and ending as a 1$\PLH$1$\PLH$2 probability vector.

\begin{figure}[h]
    \centering
    \subfloat[The 32 3$\PLH$3$\PLH$2 kernels which convolve the input image in the first convolutional layer]{\includegraphics[width=5.8cm,trim=3cm 2.8cm 3cm 3cm]{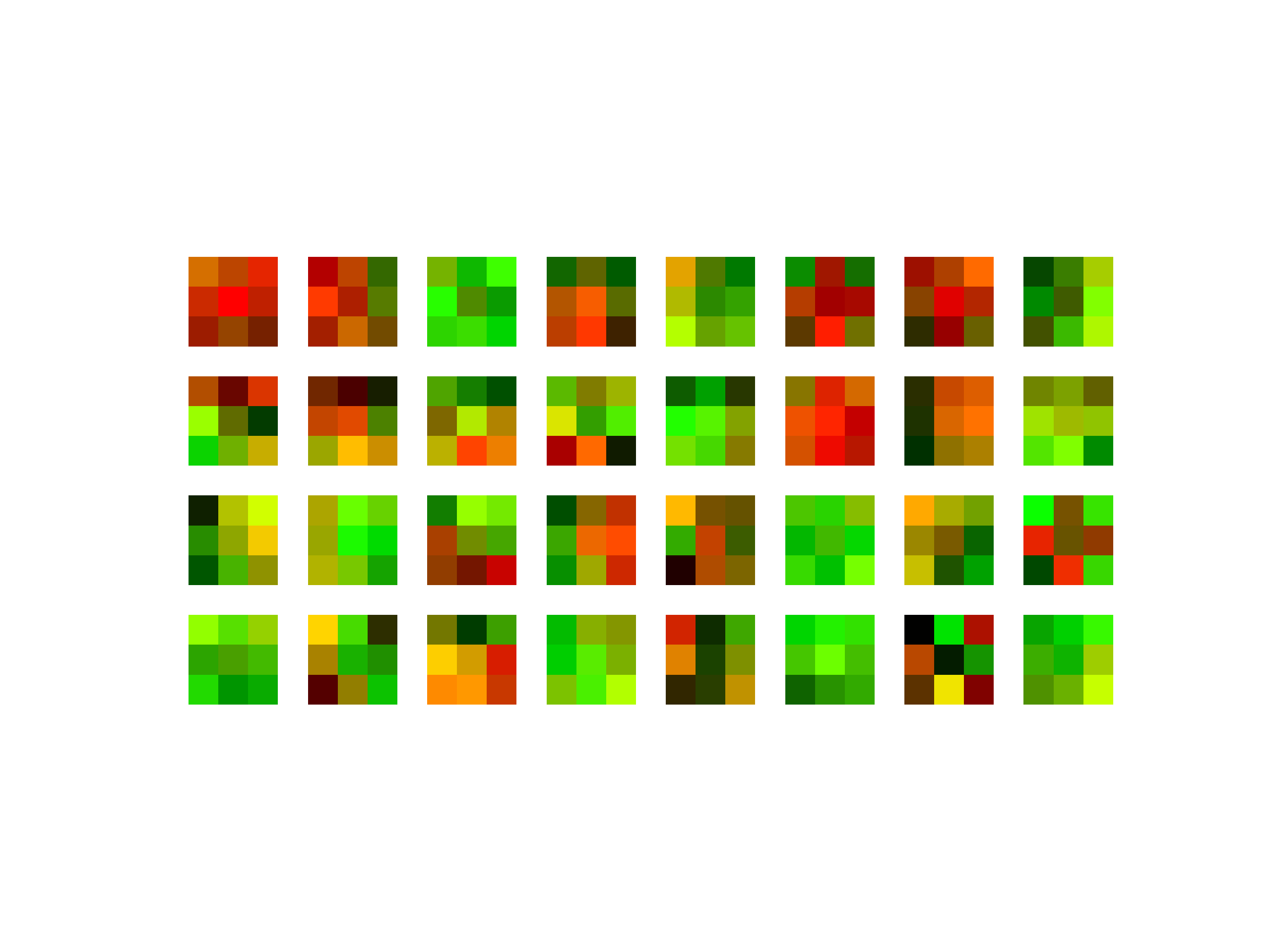} }%

    \subfloat[The 32 3$\PLH$3$\PLH$32 kernels which convolve the feature tensor in the second convolutional layer]{\includegraphics[width=6.4cm,trim=6cm 3.5cm 6.0cm 9.5cm]{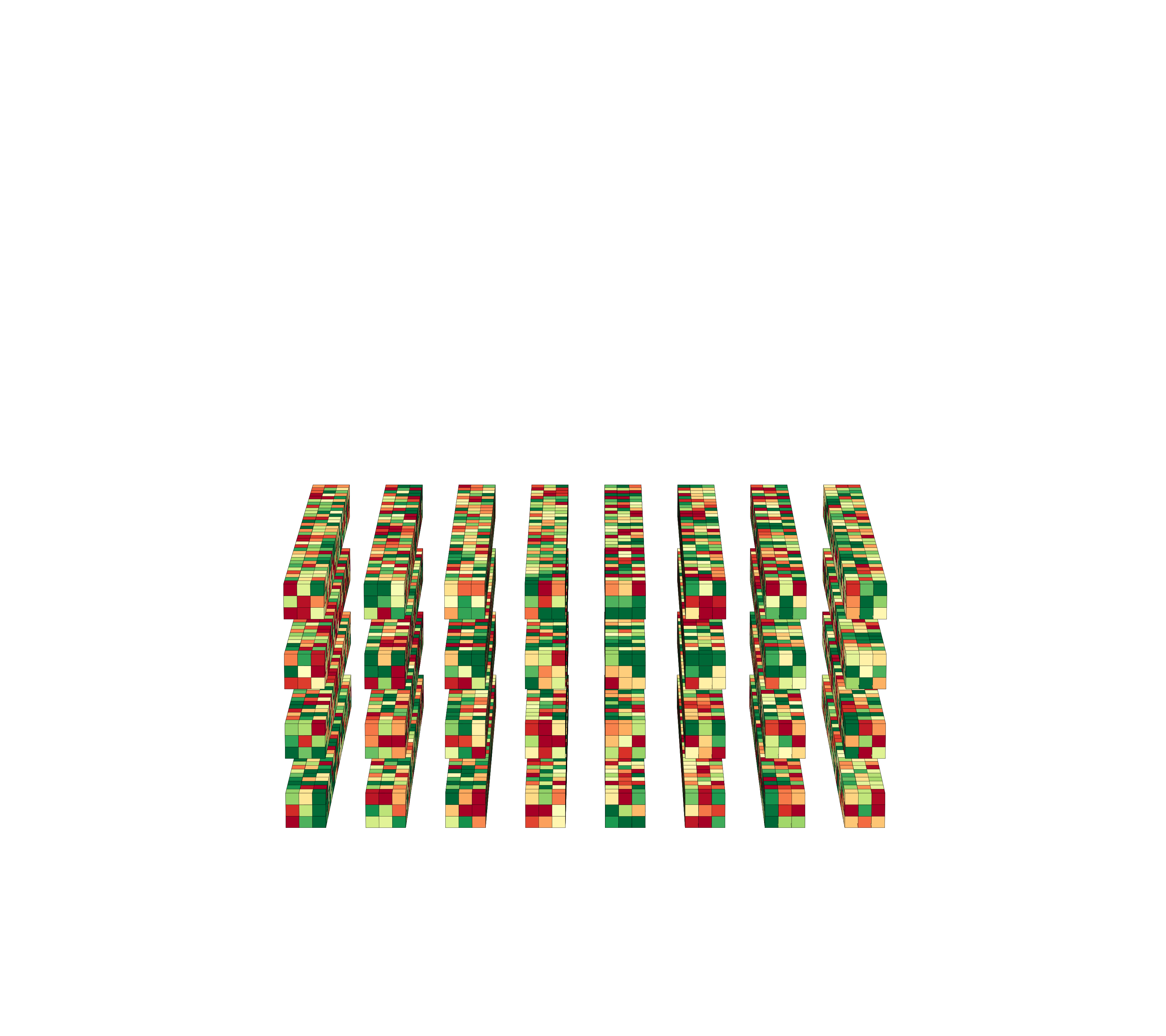} }%
    \caption{\footnotesize Representations of the first two sets of convolutional kernels learned by our network. Those in (a) are displayed using two color channels (red \& green) and those in (b) are displayed using a linear colormap which runs from green to red.}%
    \label{fig:filters}%
\end{figure}

The majority of the trainable parameters in a convolutional neural network are generally found in the convolution filters (also called convolution kernels). The first two sets of kernels from our trained network are displayed in figure \ref{fig:filters}. As one might expect, observing the values of these kernels directly provides us with little to no information; however, it is importing to bear in mind it is the action of these kernels we are attempting to understand. 

The power of the convolution operation comes from the fact that although at each spatial position the kernels act only on a small spatial area, they span the full depth of the feature tensor. This allows the different aspects (i.e. features) of our cell image held in the different channels to `communicate' with one another. As the feature tensor ascends the network and is repeatedly convolved by the various sets of learned kernels, the features that each channel represents become more complex, describing increasingly higher-level ideas about the cell image. It is what these features are and how they change as we ascend the network that we endeavour to (at least partially) understand in this investigation.

\subsection{Neuron Activations} \label{ssec:neuron_act}

\begin{figure}
\includegraphics[width=8cm ,trim=2.5cm 2.5cm 2.3cm 2.8cm]{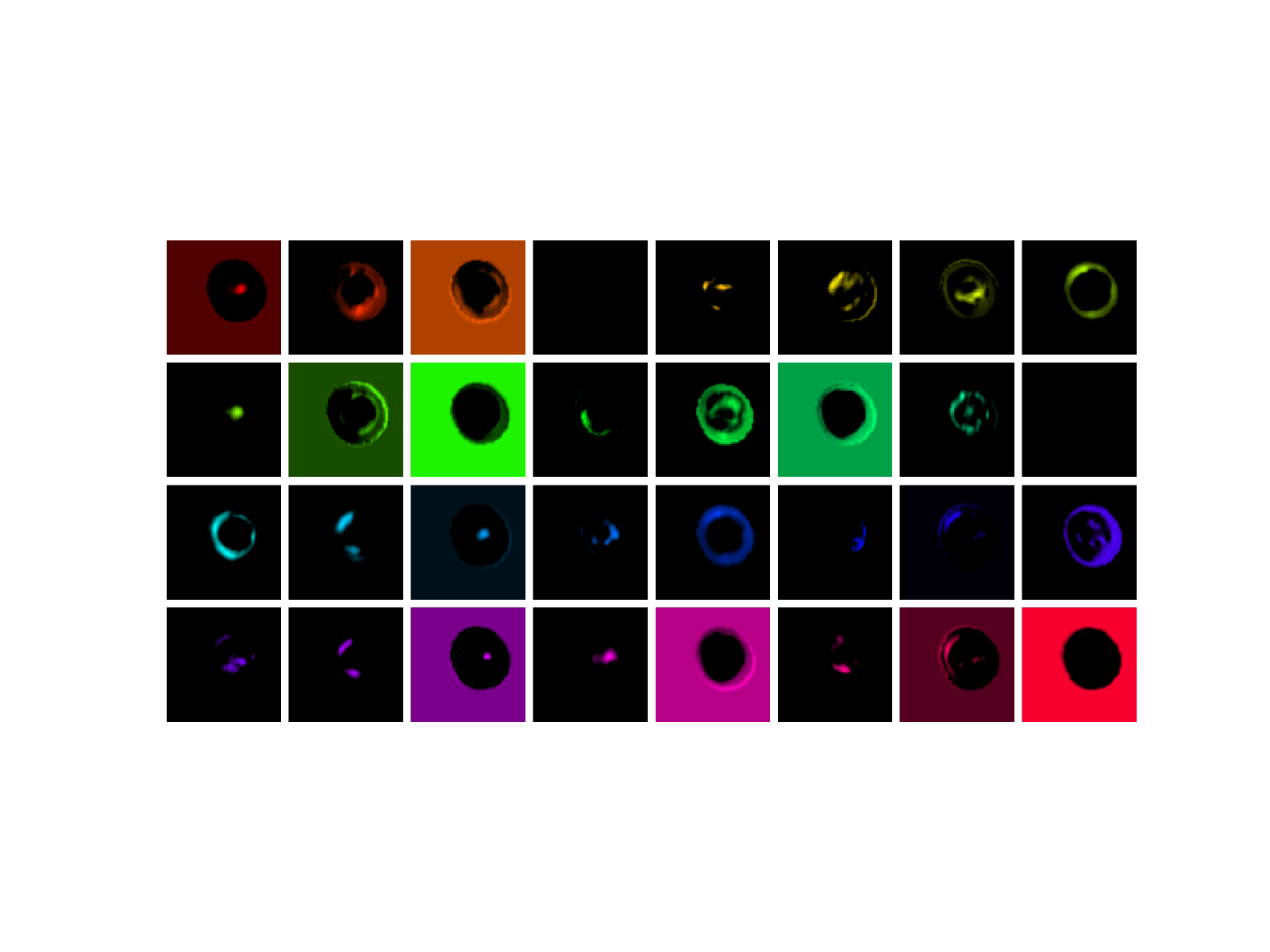}
\caption{\footnotesize Neuron activations of the 32 channels from the 2\textsuperscript{nd} convolutional layer. Each channel is displayed in a different hue and has been normalised channel-wise.}
\label{fig:act_allchan}%
\end{figure}

Directly observing neuron activations is a simplistic approach to visualising a neural network. However, it can provide us with some important information, particularly when used in conjunction with other visualisation techniques (see section \ref{ssec:interpreting}). Figure \ref{fig:act_allchan} shows the neuron activations of each channel in the 2\textsuperscript{nd} convolutional layer for the cell shown in figure \ref{fig:image_filter}, where the activations for each channel are displayed in a different hue. Viewing activation maps directly works well in the early layers, as the spatial dimensions of the feature tensor are still relatively large and the activation maps are therefore not too pixelated.

\begin{figure}
\includegraphics[width=6cm ,trim=1.2cm 0.4cm 0.8cm 0.3cm]{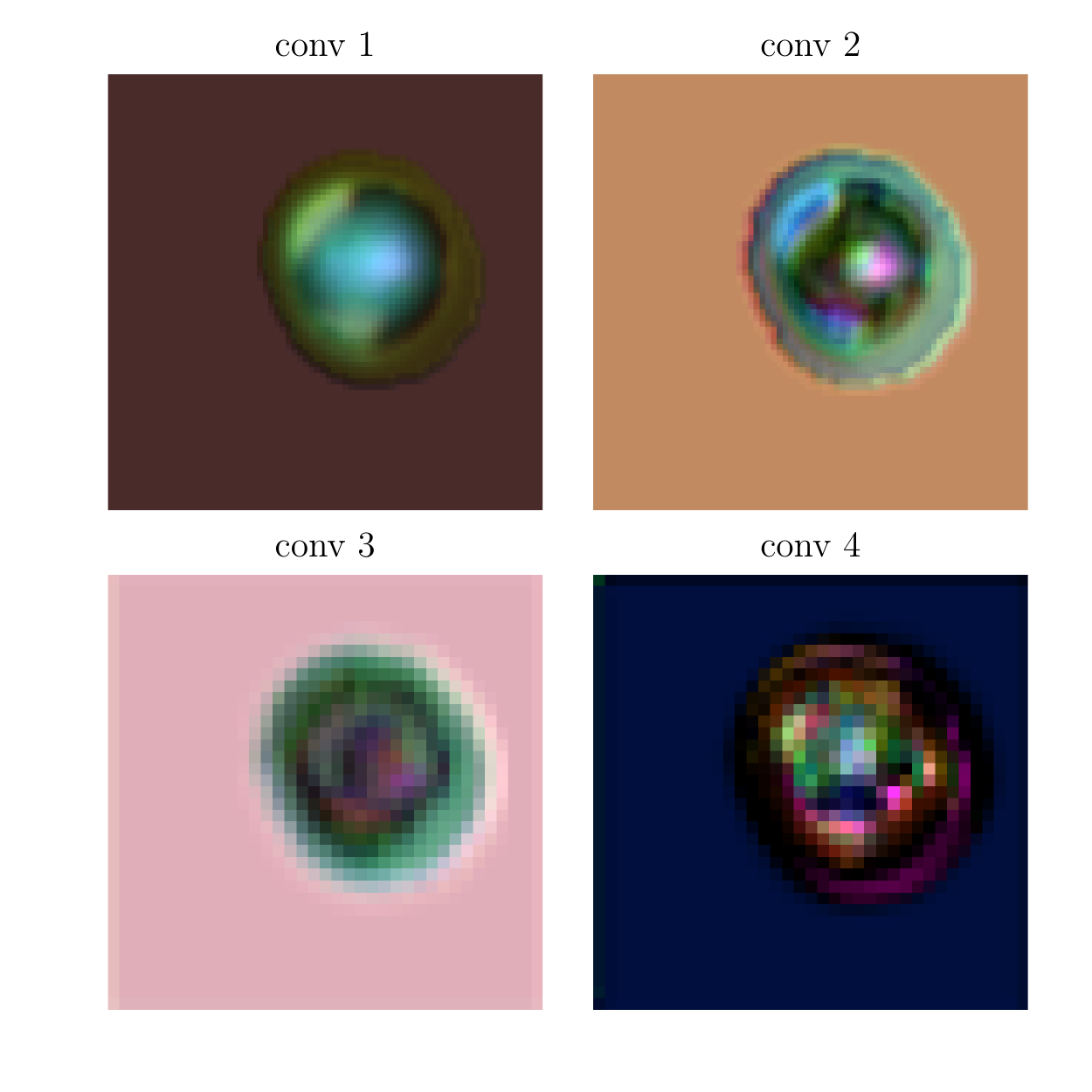}
\caption{\footnotesize Representations of the cell image (shown in fig \protect\ref{fig:image_filter}) at different network depths, made by combining the neuron activations from each channel into a single image, for the first 4 covolutional layers.}
\label{fig:act_comb_2x2}%
\end{figure}

By combining the channel activations into a single image we can form a representation of how a specific layer of the network `sees' a given cell. Figure \ref{fig:act_comb_2x2} does this for the first four convolutional layers in our network for the cell image shown in figure \ref{fig:image_filter}.

\section{Dimension Reduction Visualisations}
\label{S:2}

\begin{figure*}
\includegraphics[width=\textwidth ,trim=0.5cm 0cm 0cm 0cm]{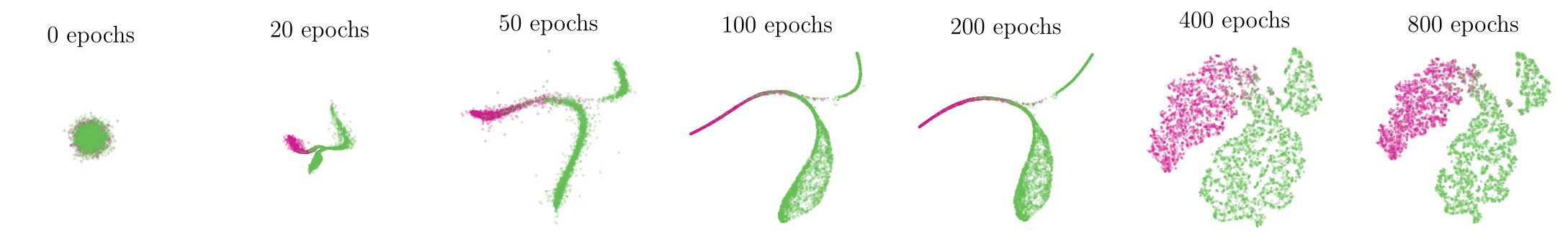}
\captionsetup{width=0.68\textwidth}
\caption{\footnotesize t-SNE embedding of the 1\textsuperscript{st} dense layer (perplexity = 30), shown for a varying number of iterations (or \textit{epochs}) of the embedding algorithm.}
\label{fig:tsne_gen}%
\end{figure*}

Dimension reduction is the process of finding a low dimensional representation of high dimensional data, while preserving underlying structures in the data. In the context of convolutional neural networks, dimension reduction is typically used to embed the high dimensional data from a dense layer into a two dimensional space. This is primarily used to help visualise how well a dataset has been separated prior to actual classification, as well as to observe structures within the dataset. Ideally, we want the different classes of data to be separated into well-defined clusters of points. 

The current standard technique for dimension reduction is t-Distributed Stochastic Neighbour Embedding (or t-SNE) \cite{t-SNE}, which dramatically improves upon the techniques that came before it, such as PCA \cite{PCA}, multidimensional scaling \cite{Multidimensional scaling} and Isomap \cite{Isomap}. Uniform Manifold Approximation and Projection (or UMAP) is another recently developed dimension reduction technique which presents several potential advantages over t-SNE \cite{UMAP}. In this investigation we focus our attention on t-SNE and UMAP. 

\subsection{t-SNE} \label{ssec:t-SNE}

t-SNE is a non-parametric mapping from a high dimensional dataset to an embedding manifold (generally a 2D manifold) that is learned iteratively. Generally, this mapping process is terminated once a stable embedded configuration is reached. Figure \ref{fig:tsne_gen} shows seven snapshots of this embedding process, with the last snapshot being the stable configuration. With some effort this embedding process can be animated; which reveals a surprisingly beautiful, organic-looking evolution.

The form of a given t-SNE embedding is controlled by two main hyperparameters: the perplexity and the learning rate. The perplexity effectively sets the number of nearest neighbours, with more dense datasets requiring higher perplexity values. It is typically set within the range 5 - 50 and changing its value can quite radically change the appearance of the plot. For an excellent interactive guide to using t-SNE, see the work by Wattenberg et al. \cite{usingtsne}. 

Interpreting a t-SNE plot is normally quite intuitive, however there are a number of important factors to bear in mind. The first is that generally the spatial area a cluster of points covers does not mean anything; what is important is that the points are clustered at all. The second is that distances between well-separated clusters of points are generally not meaningful. Following this, it is typically difficult to interpret topological information from t-SNE plots; however by observing multiple perplexities, some topological information may be inferred.

\subsection{UMAP} \label{ssec:UMAP}

UMAP is a dimension reduction algorithm that was recently developed, which for our purposes has several potential advantages over t-SNE. Firstly, UMAP is significantly faster than t-SNE, particularly for the large datasets. This makes the process of generation and parameter selection notably quicker. Secondly, according to McInnes and Healy, UMAP often performs better than t-SNE at preserving global and topological structures in the dataset, while still maintaining local neighbour relations. Lastly, in our experience, UMAP is slightly more user steerable than t-SNE and uses more intuitive hyperparameters. In UMAP, the minimum distance hyperparameter allows us to directly control how tightly the embedding compresses points together, and the $n$-neighbours hyperparameter allows us to control the degree to which the embedding preserves neighbour relations over more global structures.

UMAP embeddings can be very easily implemented using the \code{umap} function from umap-learn python library \cite{umap-learn}. This has identical usage to the \code{tsne} function from the scikit-learn library \cite{scikit} meaning we can easily use UMAP and t-SNE interchangeably in our code.

\subsection{Enhancing the Embedded Information Space} \label{ssec:dim red enhancing}

\begin{figure}[h]

    \centering
    \subfloat[Point size scaled by certainty]{{\includegraphics[width=8cm,trim=0cm 0cm 0cm 0cm]{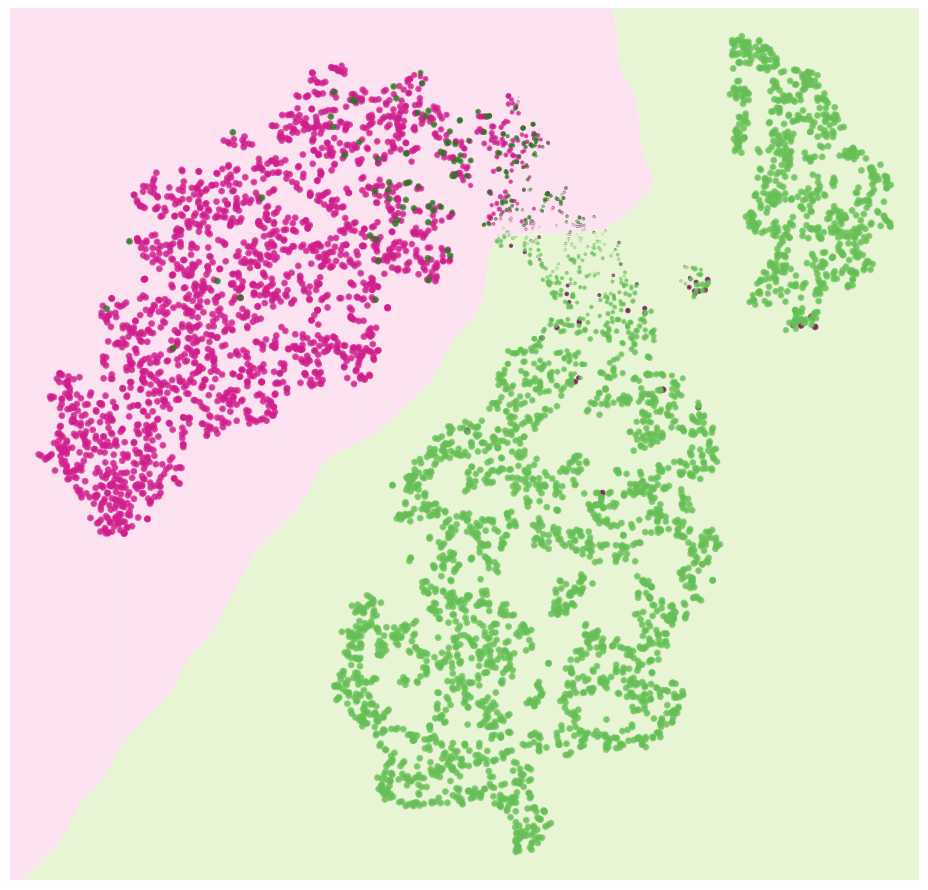} }}%

    \subfloat[Point size scaled by uncertainty]{{\includegraphics[width=8cm,trim=0cm 0cm 0cm 0cm]{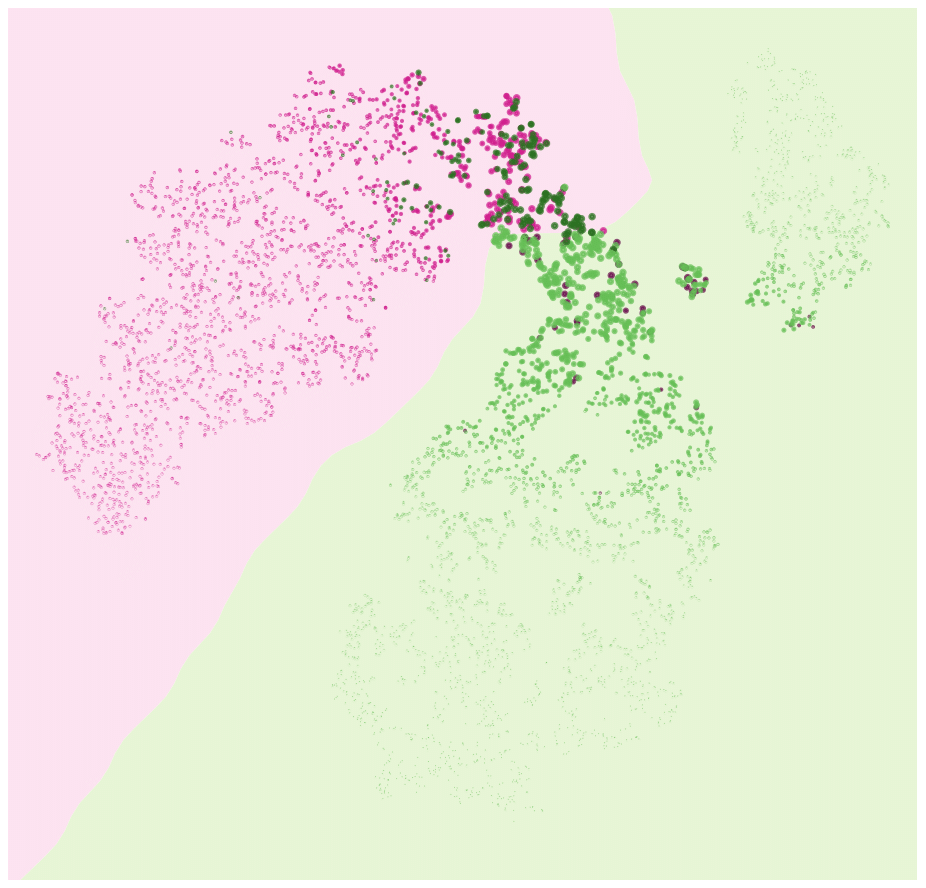} }}%
    \captionsetup{width=7.8cm}
    \caption{\footnotesize t-SNE embeddings of the 1\textsuperscript{st} dense layer (perplexity = 30), with misclassification indication, estimated decision-boundaries and certainty/uncertainty scaling.}
    \label{fig:tsne_scaling}
\end{figure}

Due to the ubiquity of dimension reduction visualisation techniques, several methods to enhance the interpretability of their information space were explored. The first and most simple addition made was to add an indication of which data-points were incorrectly classified by the network. In figure \ref{fig:tsne_scaling} this is done altering the colour of the misclassified points to a darker shade. Another technique developed was to scale the size of the data-points by the certainty or uncertainty with which they were classified. This technique is also displayed in figure \ref{fig:tsne_scaling}.

The decision boundary in the embedding space can be approximated using a nearest neighbour classification technique \cite{approx_nn}. This effectively involves producing a Voronoi diagram\footnote{A Voronoi diagram (with k = 1) is where a plane of $n$ points is partioned into polygons such that each polgon contains a single point and in each polygon the closest point in the plane is the point which the polygon contains. \cite{voronoi}} for the embedded data-points and colouring the polygons according to each data-point’s predicted class. In figure \ref{fig:tsne_scaling}, this was implemented using the \code{KNeighborsClassifier} from the scikit-learn library \cite{scikit} (using k = 3 to reduce edge noise) and smoothing the resulting boundary with a Gaussian filter. Other more experimental approaches exist which attempt to better approximate the true decision boundary \cite{exp_nn}, however these were not investigated here. 

\begin{figure*}
\includegraphics[width=18cm ,trim=0cm 0cm 0cm 3cm]{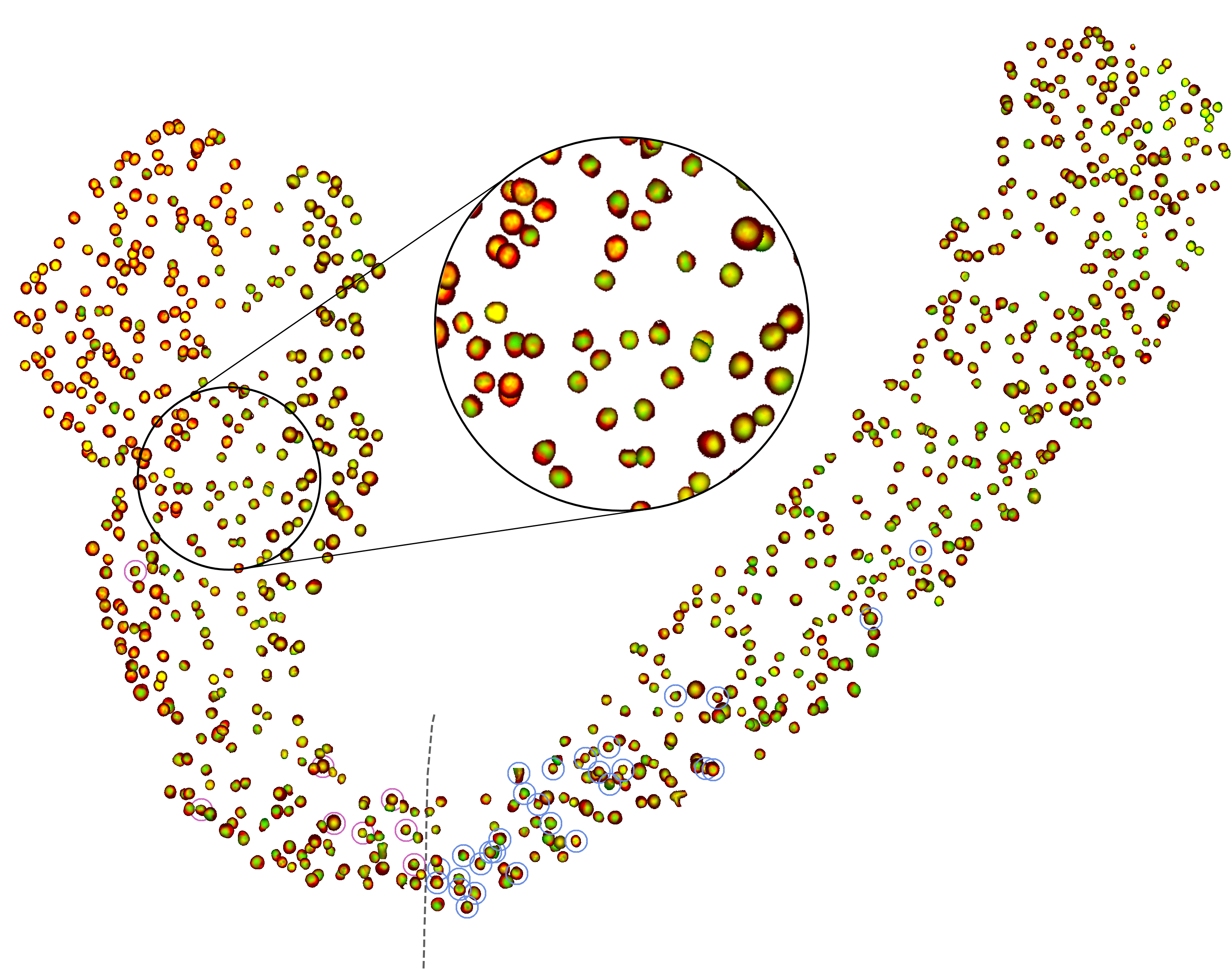}
\caption{\footnotesize UMAP embedding of the output of the 1\textsuperscript{st} dense layer for 450 lymphocyte cell images and 450 neutrophil cell images. For this UMAP embedding a high minimum distance value (see section \protect\ref{ssec:UMAP}) was used to reduce point overlap and improve visibility, as each point is represented by its corresponding cell image with the background removed. The salient section of the estimated decision boundary has also been plotted with a dashed line. Cells to the left of this boundary were classified as neutrophil cells and cells to the right were classified as lymphocyte cells. Neutrophil cells which were misclassified as lymphocyte cells are circled in pink, and lymphocyte cells which were misclassified as neutrophil cells are circled in blue.}
\label{fig:tsne_cells}%
\end{figure*}

A method that we have developed to better understand the `geography' of the embedded space is to represent each point with its corresponding cell image. This is done in figure \ref{fig:tsne_cells}, which reveals several well defined groups of similar cell morphologies within the dataset. 

\subsection{Grid Mapping} \label{ssec:dim red grid}

\begin{figure*}
\includegraphics[width=\textwidth ,trim=0cm 0cm 0cm 0cm]{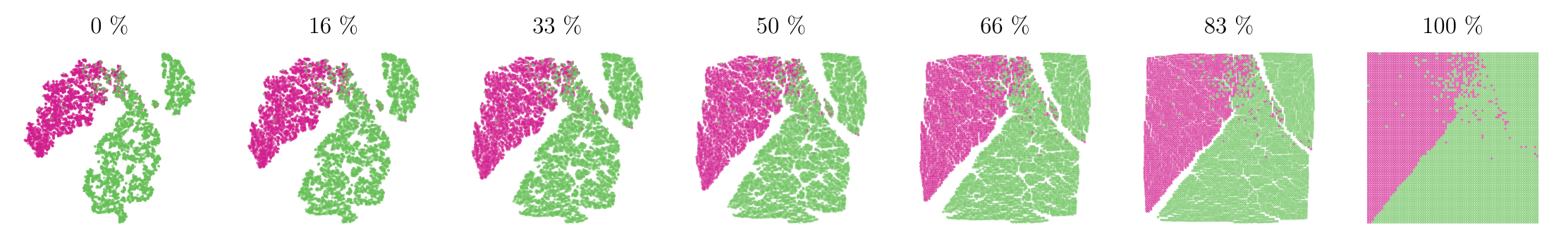}
\caption{\footnotesize A t-SNE embedding of the 1\textsuperscript{st} dense layer before and after grid-mapping, showing five intermediate stages. The percentage indicates the fraction which each point has been moved from its unmapped point to its mapped point.}
\label{fig:tsne_grid_series}%
\end{figure*}

For several reasons which will shortly become apparent, it can be useful to map the embedded datapoints to a grid. Using the Jonker-Volgenant algorithm \cite{gridmap} to calculate the shortest augmenting path, the desired linear mapping can be calculated relatively quickly. The Linear Assignment Problem solver library was used to produce the examples below \cite{lapjv}. Figure \ref{fig:tsne_grid_series} illustrates this process, showing a t-SNE embedding being gradually mapped to a grid.

\begin{figure}
\includegraphics[width=7.5cm ,trim=0cm -0.2cm 0cm 0cm]{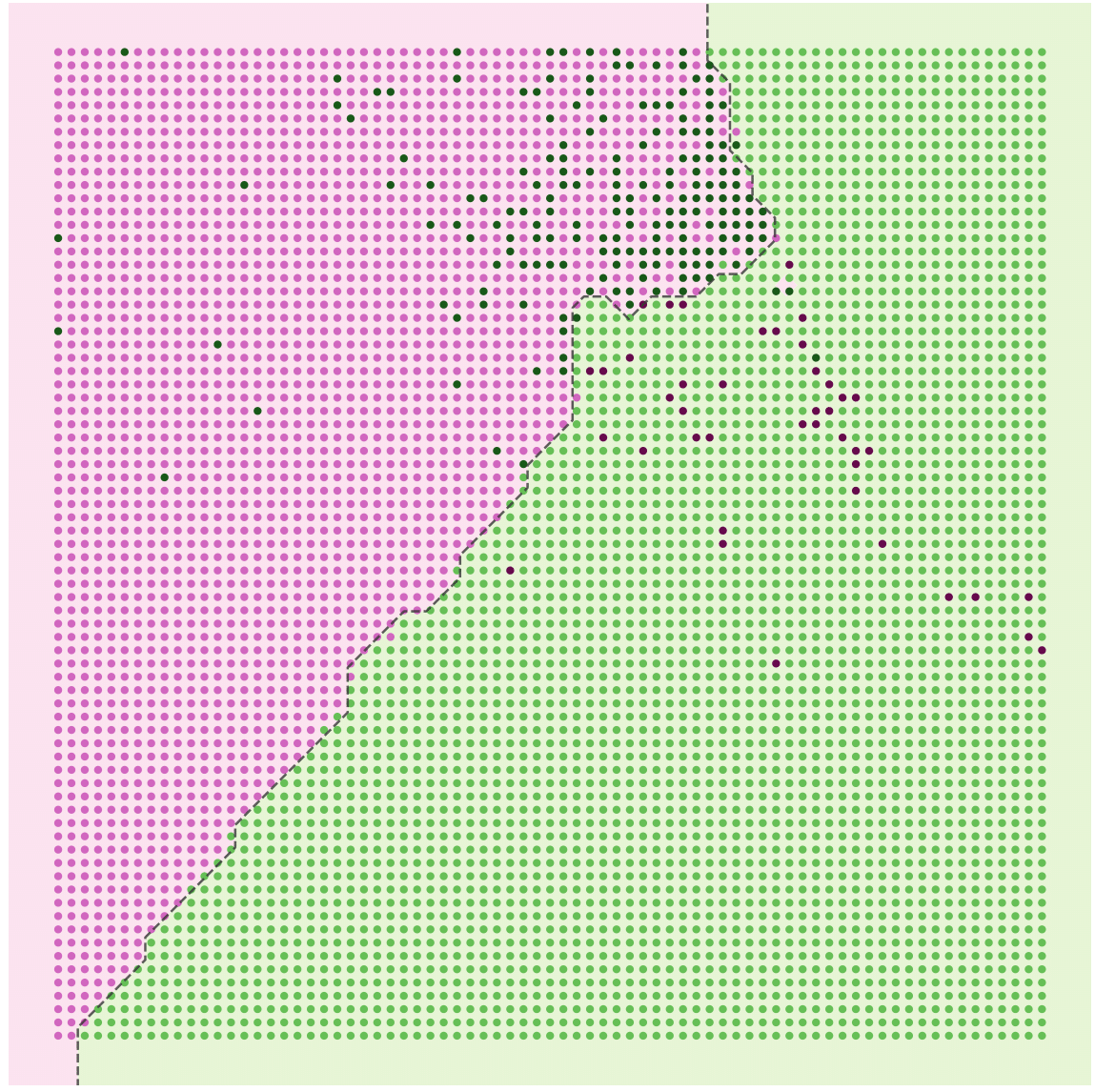}
\captionsetup{width=8.5cm}
\caption{\footnotesize A fully grid-mapped t-SNE embedding of the 1\textsuperscript{st} dense layer with misclassification indication. The decision boundary has also been estimated for the grid-mapped points and is highlighted by a dashed grey line.}
\label{fig:tsne_grid_boundary}%
\end{figure}

Using grid mapping can give us a different perspective on the embedded space. With every data-point equally visible, the relative distributions of the two classes about the decision boundary can be more easily seen; see figure \ref{fig:tsne_grid_boundary}. An estimate of the decision boundary in the grid-mapped space is also shown in this figure, calculated using the same approach as in section \ref{ssec:dim red enhancing}.

To assist in the generation and exploration of different embeddings, and the associated techniques discussed here; a simple HTML-based interface was developed using a Python visualisation library called Bokeh \cite{bokeh}. This interface includes classification certainty/uncertainty scaling, misclassification highlighting and grid mapping (from 0 to 100\%). Additional features include cell image viewing by tapping data-points and attribution pie charts which dynamically update based on user selection. 

\subsection{Flattened Feature Tensor Embedding} \label{ssec:dim red flat}

\begin{figure}
\includegraphics[width=7.81cm ,trim=0cm 0cm 0cm 0cm]{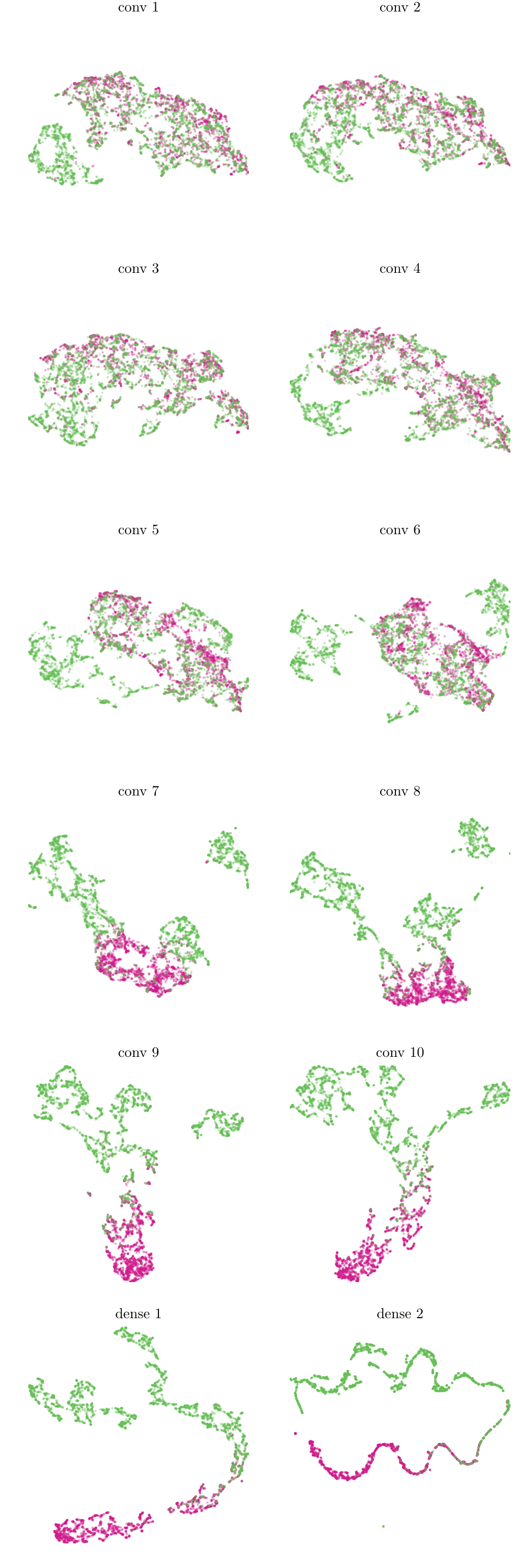}
\caption{\footnotesize UMAP embeddings of the flattened feature tensor for the 10 convolutional layers and first 2 dense layers, for 2180 lymphocyte images (pink) and 3484 neutrophil images (green). UMAP embedding used $n$ neighbours = 15 and a minimum distance of 0.1.}
\label{fig:tsne_flatten}%
\end{figure}

Conventionally, embedding techniques are used only on the dense layers, where the feature tensor has been reduced to a high-dimensional column vector. However, by intercepting  the feature tensor prior to the dense layers and flattening it into a column vector, we can embed a dataset from any point in the network. Doing this for each layer of the network provides us with a method to directly visualise the development of class separation, see figure \ref{fig:tsne_flatten}. Note that these embeddings where produced using UMAP due to its superior speed, and each embedding was produced using the same seed, allowing us to better observe how structures in the dataset evolve as we move deeper into the network. 

These techniques may also be able to provide us with some other pieces of information. For example in figure \ref{fig:tsne_flatten}, in the embedding of the first convolutional layer (top left) we can observe a cluster of neutrophil cells (shown in green) already fairly well separated from the rest of the dataset. This cluster likely represents a group of cell images we would be able to distinguish `by eye'. Additionally, this technique may be able to inform us to the presence of redundancy in the architecture of our model. If there are redundant layers in our network, then class separation will likely be very high prior to reaching the dense layers. In fact, this is likely the case for our model, where from the 7\textsuperscript{th} layer onwards we observe high levels of class separation.

\section{Feature Visualization by Optimization}
\label{S:3}

The fundamental idea of feature visualisation by optimisation is simple and has existed for quite some time \cite{Erhan}. At its fundamental level it is the process of finding patterns, textures and shapes which maximize the activation of a neuron or combination of neurons (i.e. units of the feature tensor). The reasoning is that by finding images to which a group of neurons is responding maximally, we are finding a good first-order representation of what those neurons have learned to look for. As such, these patterns, textures and shapes can be interpreted as the learned features of our deep neural network. However, it is only until recently through the development of various regularisation and preconditioning techniques \cite{feature-vis, deepvis, invert, fooled, inceptionism, googlenet, bilateral, deepdreaming, synthgen, plugnplay, intraclass} that the images generated via this optimisation method have become recognizable and interpretable on a human level.

Adopting the formalism of Erhan et al. \cite{Erhan}, let $\theta$ denote our trained neural network and let $h_{ij}(\theta,x)$ be the activation of a unit $i$ in a layer $j$ in the network; where $h_{ij}$ is a function of both $\theta$ and the input sample $x$. Our objective is therefore to find the bounded value of $x$ which maximises $h_{ij}(\theta,x)$ (for our fixed $\theta$). This is a non-convex optimisation problem; meaning it does not have globally optimal solution. We can however try to find, or at least approach, a local minimum. This is done by performing gradient ascent in the input space; computing the gradient in $h_{ij}(\theta,x)$ and shifting $x$ in the direction of this gradient to maximise $h_{ij}$. As with gradient decent, the size of these shifts (the learning rate) and the stopping criterion are hyperparameters we must choose. However, as mentioned previously, it is only through the recent development of various conditioning techniques that we have been able to approach minima of the input space which are semantically meaningful and interpretable to us as humans. We will refer to the images we generate using this method as \textit{feature images} from here on.

\subsection{Preconditioning \& Regularisation} \label{ssec:regularization}

An adversarial example is an image which has been changed in an imperceptible way such that it becomes incorrectly diagnosed by a network; yet without this change would be correctly diagnosed with high confidence \cite{intriguing}. Related to this, it is also possible to generate images which are unrecognisable to the human eye, but which the network classifies with extremely high confidence \cite{fooled}. Termed ‘fooling images’, the main challenge in using feature visualisation by optimisation is avoiding generating such images which fool the neurons of our network into maximally activating. If we optimise for long enough, we generally start to see something vaguely interpretable; however unregularized images are characterised by large amounts of noise and high-frequency patterns. Often one can observe the generation of checkerboard-type artefacts, which have been linked to the deconvolution calculation that is done when computing the gradient in $h_{ij}$ at convolutional layers \cite{checkerboard}. Less organized high-frequency patterns can similarly be linked to max pooling operations \cite{geodesics}.

Regularisation involves placing some restrictions on the optimisation process, or adding information in the form of priors. In this investigation we restrict our focus to ‘weak’ regularisation methods which do not use learned priors to generate feature images. Weak regularisation methods instead use simple heuristics to improve interpretability. The methods we use here fall into two main categories: frequency penalisation and transformation robustness. 

Frequency penalisation is used to directly reduce unwanted high frequencies in the feature images.  Various approaches to do this exist such as blurring the image after each optimisation step \cite{fooled} and directly penalising variation between neighbouring pixels \cite{invert}. Similar techniques have also been applied to the gradient to remove high frequencies before they manifest in the image \cite{googlenet, deepdreaming}. These techniques do however discourage genuine high-frequency features such as edges. Bilateral filters, which are non-linear, edge-preserving, noise-reducing smoothing filters, could potentially be used to overcome this problem \cite{bilateral}, however these were not investigated here.

Transformation robustness refers to the approach of finding feature images which strongly activate a given objective even if the images are slightly transformed. This is done by rotating, scaling, or stochastically jittering an image by small amounts before each optimisation step \cite{inceptionism}. This set of techniques is generally highly effective, particularly when combined the aforementioned techniques \cite{googlenet, deepdreaming}, and was used to produce the feature images shown in this paper.

Formally, preconditioning techniques are techniques used in optimisation processes which apply a transformation to condition a problem into a form that is more suitable for numerical solving methods. Here, preconditioning is used to optimise activation objectives in a space with a different parameterisation or distance metric. One example of a highly effective preconditioning technique is to optimise in the Fourier basis. Doing this spatially decorrelates the image data and is used in the generation of all the images shown in this paper. With slightly more effort, colour channels can also be decorrelated. This is done by measuring the colour-correlation in some image-set, then using this measurement and a Cholesky decomposition to decorrelate colours in the genration process. Colour decorrelation was not however used here. See the work by Olah et al. for a more detailed explanation of why these techniques are effective \cite{feature-vis}.

An approach that was not explored in this investigation, but certainly presents an area for further work, is using learned priors. This technique is a form of strong regularisation and number of different approaches have recently been developed \cite{inceptionism, synthgen, plugnplay, intraclass}. For example, one approach is to use a variational autoencoder or generative adversarial network to map images into a learned latent space. Feature images could then be generated by decoding a feature tensor optimised within this latent space.

The appeal of using learned priors is that they are able to generate photorealistic images which are highly recognisable \cite{synthgen, plugnplay}. A potential drawback of these approaches is that we are not necessarily able to easily decouple what feature information comes from the prior, and what comes from the objective being optimised. Whether this is actually a point for concern is questionable, as the prior could be viewed as simply providing us with more realistic and semantically meaningful features.

For an excellent series of interactive visualisations and more detailed analysis of many of the regularisation and precoditioning methods described here, see the work by Olah et al. \cite{feature-vis}.

\subsection{Optimsation Objectives} \label{ssec:objectives}

Although it is possible to optimise for the activation of a single neuron of our network (or in other words the value of a single unit of the feature tensor), it is far more informative to optimise for the activation of combinations of neurons. One of the most straightforward ways we can break up the feature tensor is channel-wise; generating feature images which maximise the total activation of a specific channel (in other words the sum of values in a depth-wise unit slice of the feature tensor). Related to this, we can also optimise for multiple channels simultaneously, and weight the degree to which we optimise for the activation of each by different amounts. This technique is used later in section \ref{ssec:Visualising Clusters}.

In generating the feature images shown in this paper we have made significant usage of a TensorFlow-based library called Lucid \cite{lucid}. This library is an exceptional collection of tools and infrastructure for neural network interpretability and feature visualisation. It has been essential to the work presented here and at the time of writing represents the state-of-the-art in feature visualisation techniques.

\begin{figure}
\includegraphics[width=8cm ,trim=0cm 0cm 0cm 0cm]{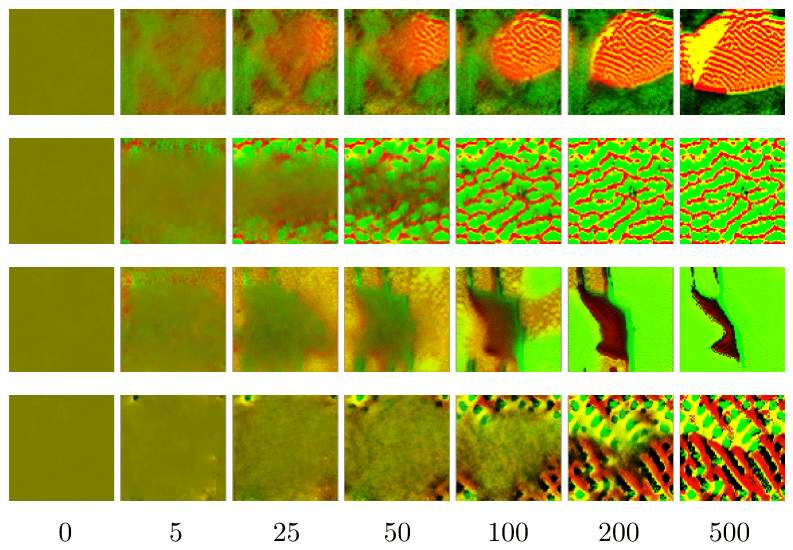}
\captionsetup{width=8.8cm}
\caption{\footnotesize Generation process of four single-channel feature images from the 4\textsuperscript{th} convolutional layer. The numbers underneath each column indicate the number of optimisation steps the image has been through. These images were generated using a Adam optimiser with a learning rate of 0.05, transformation robustness and spatial decorrelation.}
\label{fig:feature_gen}%
\end{figure}

Figure \ref{fig:feature_gen} shows the generation process of four single-channel feature images from the 4\textsuperscript{th} convolutional layer. Notice how some of the images `develop' faster than others, and how structures in the image seem to form from the outside in. However, the main take away from this figure is seem in the final column of images. After too many optimisation steps the images appear `blown out' and loose any subtle definition. As such, we must be careful not to over-optimise when generating feature images. It is also noted that in order to reach a similar level of development, feature images generated for layers deeper in the network generally require a larger number of optimisation steps. 

The fine tuning of the various hyperparameters used to generate feature images can be a somewhat tedious process. As such, another simple HTML-based interface was developed in Bokeh \cite{bokeh} to speed up this process and assist in the exploration of the various objectives, regularisation methods and optimisation parameters at our disposal. 

\subsection{Interpreting Feature Images} \label{ssec:interpreting}

As described at the start of this section, as a feature tensor ascends the network, the features that each channel represent become increasing more complex; describing progressively higher-level ideas about the cell image. Our hope is that by observing how the generated feature images change as we ascend the network, we will be able to observe the how the complexity in these ideas develops.

One of the first problems that is encountered when attempting to interpret feature images is that it is not clear which part of the generated image constitutes the ‘feature’ that the respective channel has learned to look for, and which parts are simply artefacts of the generation process. To extract the salient information from a single-channel feature image, we have developed the approach of passing the image to the network and extracting the neuron activations of the feature tensor channel the image was optimised to maximally activate. This neuron activation map is then upscaled (using Lanczos interpolation) and used to modulate the brightness of the feature image. We refer to this process as \textit{activation filtering}. The hope is that this process will reveal to us the ‘true features’ in the feature images.

We can qualify in a semi-quantitative way whether information has been lost in this filtering process by passing the filtered image to the neural network and observing the activation map of its associated channel. If the salient information is preserved in the filtering process, then the activation map of the filtered and unfiltered images should be very similar (with certain exceptions). This therefore provides us with a feedback mechanism which we can use to adjust the details of the filtering to better preserve the salient information.

\begin{figure}
\includegraphics[width=9.53cm ,trim=0.1cm 0.6cm 0cm 0.06cm]{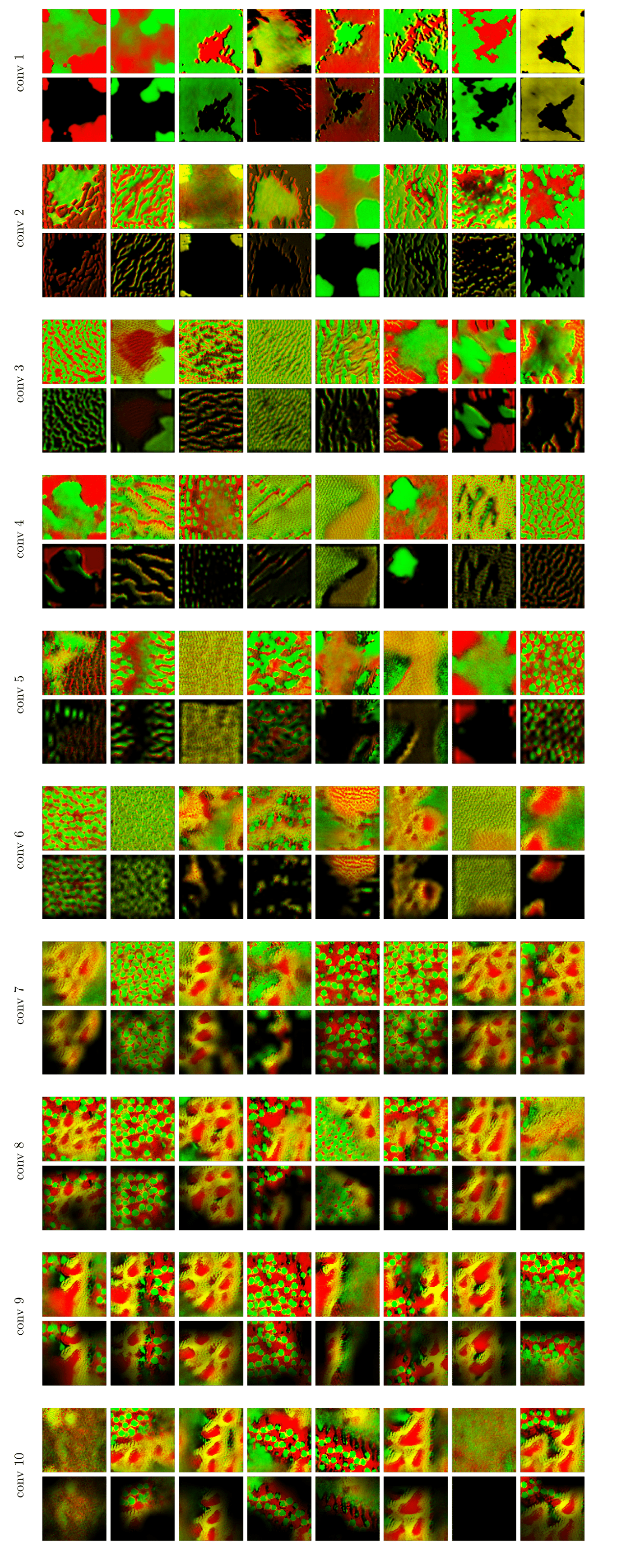}
\captionsetup{width=8cm}
\caption{\footnotesize A selection of single-channel feature images generated for each convolutional layer of the network. For each layer, the activation-filtered images are shown below their unfiltered counterpart.}
\label{fig:feature_ims_layers}%
\end{figure}

Figure \ref{fig:feature_ims_layers} shows single-channel feature images generated for the 10 convolutional layers of our network. The images shown represent only a small subset of the full set of feature images for each layer, however were hand selected to show the diversity within each layer. Each feature image is also shown before and after activation filtering.

\begin{figure}[h]
    \centering
    \subfloat[Unfiltered]{{\includegraphics[width=8.2cm,trim=0cm 0cm 0.4cm 0cm]{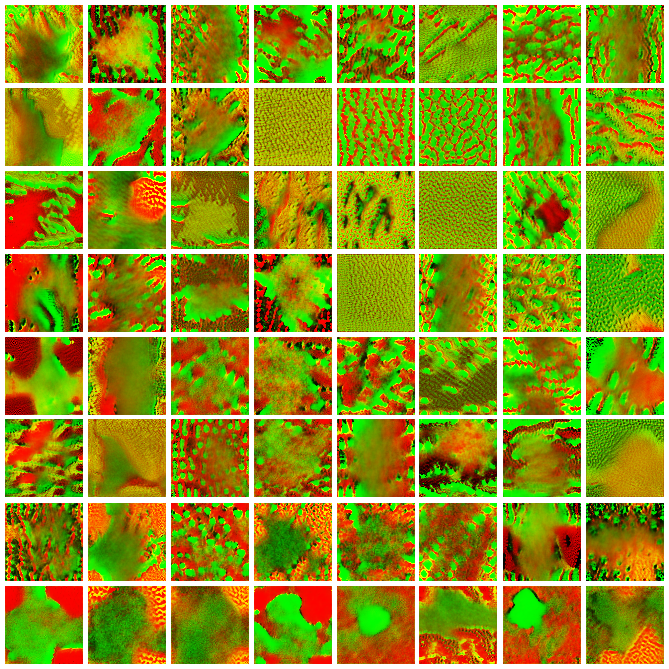} }}%
    
    \subfloat[Using activation filtering]{{\includegraphics[width=8.2cm,trim=0cm 0cm 0.4cm 0cm]{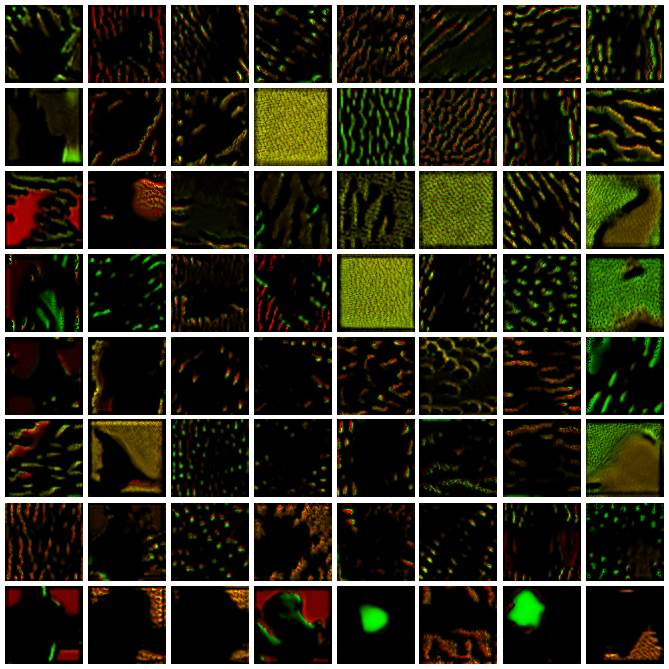} }}%
    \captionsetup{width=8cm}
	\caption{\footnotesize All 64 single-channel feature images of the 4\textsuperscript{th} convolutional layer. The images have been spatially positioned by grid-mapping a UMAP embedding of the flattened 4\textsuperscript{th} convolutional layer feature tensors of the unfiltered feature images.}
	\label{fig:feature_ims_UMAP}%
\end{figure}

By studying the images displayed in this figure, we can start to build up a general picture of how feature complexity develops. The first couple layers appear to mostly focus on block colours and edges; followed by textures and patterns, then shapes in later layers. However, beyond the 6\textsuperscript{th} convolutional layer, the feature images become more difficult to interpret (and a small number of similar patterns appear to dominate the majority of feature images). To understand why this is the case, consider an image classification network with two more familiar classes: cat and dog. As we move up through the layers of this network we expect it to look at increasing complex aspects of the input images. First edges, then textures, then patterns and shapes, then more complex ideas like facial structure and ear ‘floppiness’. In its final layers we expect the network to focus on the high-level features of cats and dogs which make them different. But what does the difference between a cat and a dog look like? When we generate feature images for these layers it is this difference we are visualising. However, the abstract nature of the difference between these two similar, but distinct classes, means the generated features images may be fundamentally undecipherable to us. This is precisely what we observe in the deeper layers of our cell classification network.

In order to try to understand how the features of a given layer relate to one another, we have used the embedding techniques described in section \ref{S:2} to position the feature images in a grid, where similar images (according to the network and the embedding) are placed next to each other. This was done by embedding the flattened feature tensors of the unfiltered feature images into a 2D space, then mapping this embedding to a grid. Figure \ref{fig:feature_ims_UMAP} does this for the 4\textsuperscript{th} convolutional layer, where the images displayed in (a) are unfiltered, and in (b) use activation filtering.

\begin{figure}
\includegraphics[width=8.9cm ,trim=0.2cm 0.2cm 0cm 1cm]{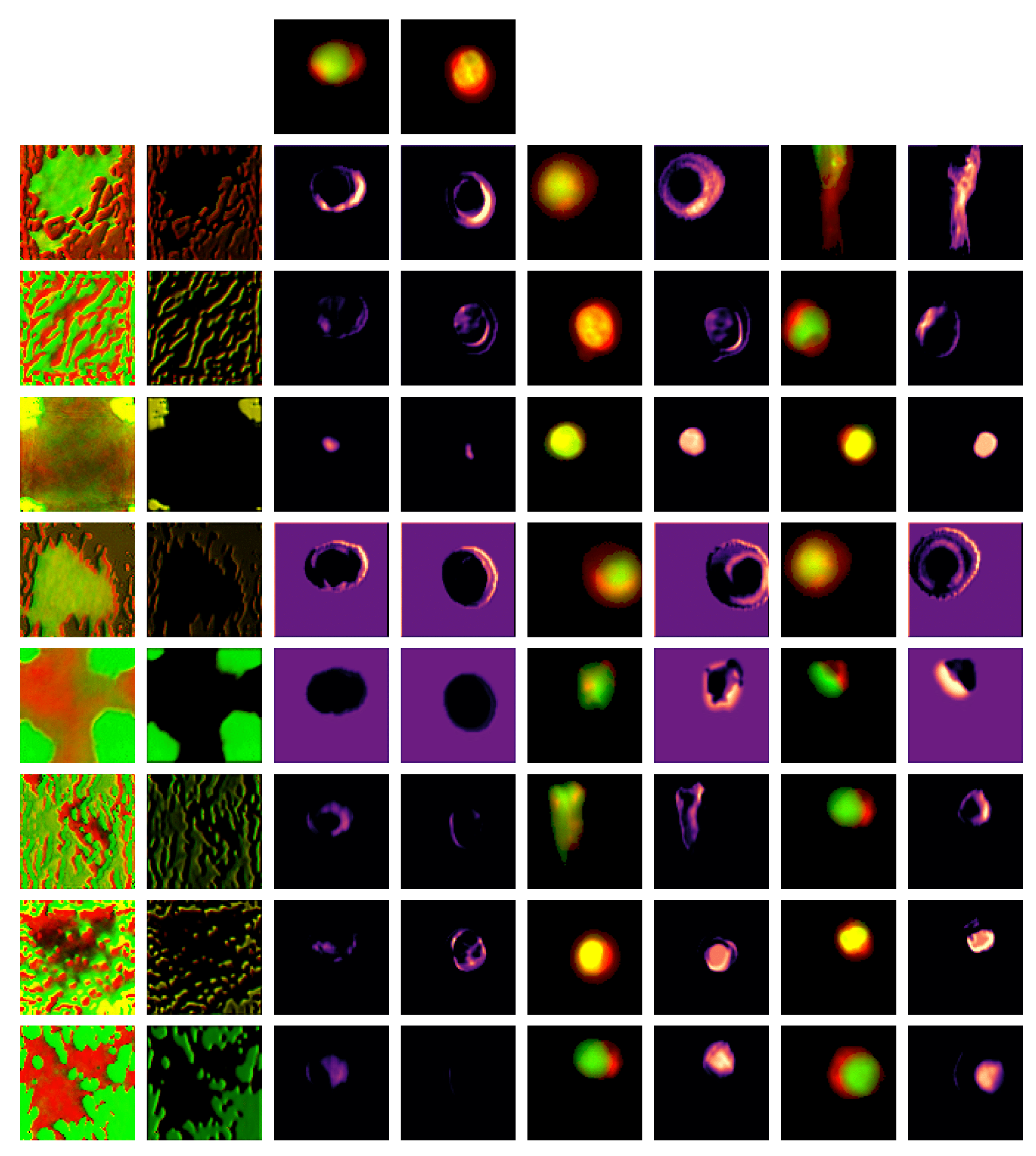}
\captionsetup{width=8.6cm}
\caption{\footnotesize Pairing feature images, activation maps and maximal images. The first two columns show 8 unfiltered and filtered single-channel feature images from the 2\textsuperscript{nd} convolutional layer (also seen in figure \protect\ref{fig:feature_ims_layers}). The next two columns show the activation maps of the respective channels for the cell images shown at the top of each column. The 5\textsuperscript{th} row shows the images from the dataset which maximally activate the respective channels; and the 6\textsuperscript{th} row shows these image's activation maps for the respective channels. The next two columns do the same for the second most maximally activating images.}
\label{fig:hue_activations_max}%
\end{figure}

We can supplement the information provided by feature images with activation maps, and by finding the images within the image-set which maximise channel activation. Figure \ref{fig:hue_activations_max} combines these different information streams and provides us with a method for understanding what a channel is looking for and how this manifests itself in the cell images.

\subsection{Feature Tensor Factorisation} \label{ssec:neuron_groups}

\begin{figure}
\includegraphics[width=8.6cm ,trim=0cm -0.3cm 0cm 0cm]{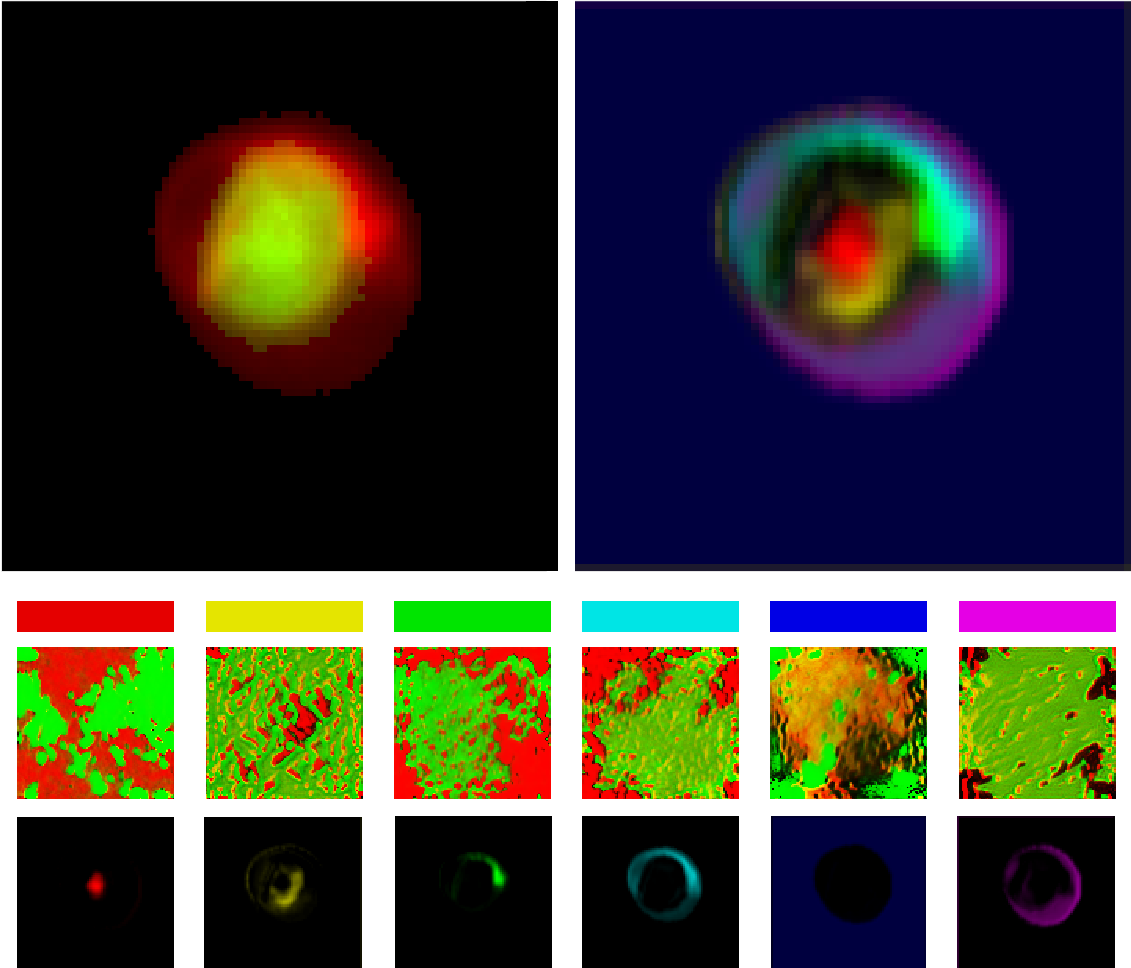}
\caption{\footnotesize Factorised feature tensor and associated feature images. The top right image shows a representation of the feature tensor from the 2\textsuperscript{nd} convolutional layer of the cell image on the left. The feature tensor has been factorised into 6 neuron groups using non-negative matrix factorisation and the activations from each group are shown in a different hue. Feature images which have been generated to maximise the total activation of each neuron group are displayed in the 2\textsuperscript{nd} row. The spatial activations of each group are shown individually in the bottom row in their respective hues.}
\label{fig:neurongroups}%
\end{figure}

An approach we can use in order to better understand a group of single-channel feature images which are similar in appearance is to factorise the feature tensor of a given cell image into $n$ groups, then generate feature images which maximally activate each of these groups of neurons. This can be useful as it is a less fine-grained approach compared to looking at single-channel feature images. Here we follow the approach used by Olah et al. \cite{building-blocks}, using non-negative matrix factorization, which was used to produce figure \ref{fig:neurongroups}.

\subsection{Visualising Embedded Clusters} \label{ssec:Visualising Clusters}

In the low dimensional embeddings of section \ref{S:2}, datapoints often cluster into fairly well defined groups. Using a clustering algorithm such as DBSCAN (Density-Based Spatial Clustering of Applications with Noise) \cite{DBSCAN}, these clusters can be identified. By analysing the properties of these clusters and using feature visualisation by optimisation, we can generate a feature image to visualise each cluster, and use these images to elucidate structures in the embedded space. The DBSCAN clustering algorithm was chosen for number of reasons. Firstly, it does not require us to specify the number of clusters we want to identify, but rather finds the number generatively according to the hyperparameters we provide it (which offers us some soft control). Secondly, in contrast to many other clustering algorithms, it is able to find arbitrarily shaped clusters by making few assumptions about cluster properties. Lastly, it handles outliers well and will identify data-points which do not fit within any of the identified clusters.

\begin{figure}
    \centering
\subfloat[3\textsuperscript{rd} convolutional layer embedding]{{\includegraphics[width=8.5cm ,trim=0.5cm 1.15cm 0.5cm 1cm]{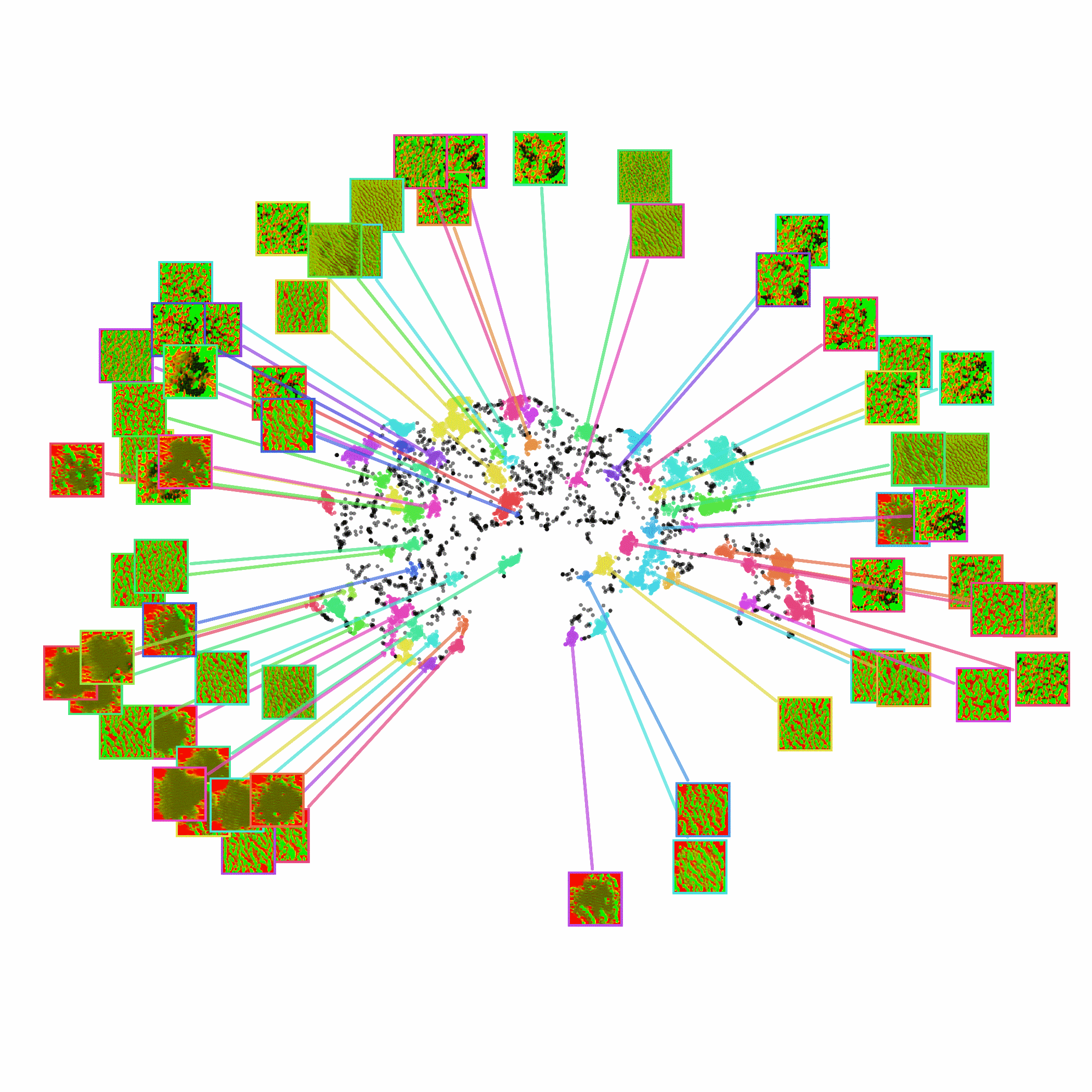} }}

\subfloat[9\textsuperscript{th} convolutional layer embedding]{{ \includegraphics[width=8.4cm ,trim=0.8cm 0.25cm 0.5cm 0.2cm]{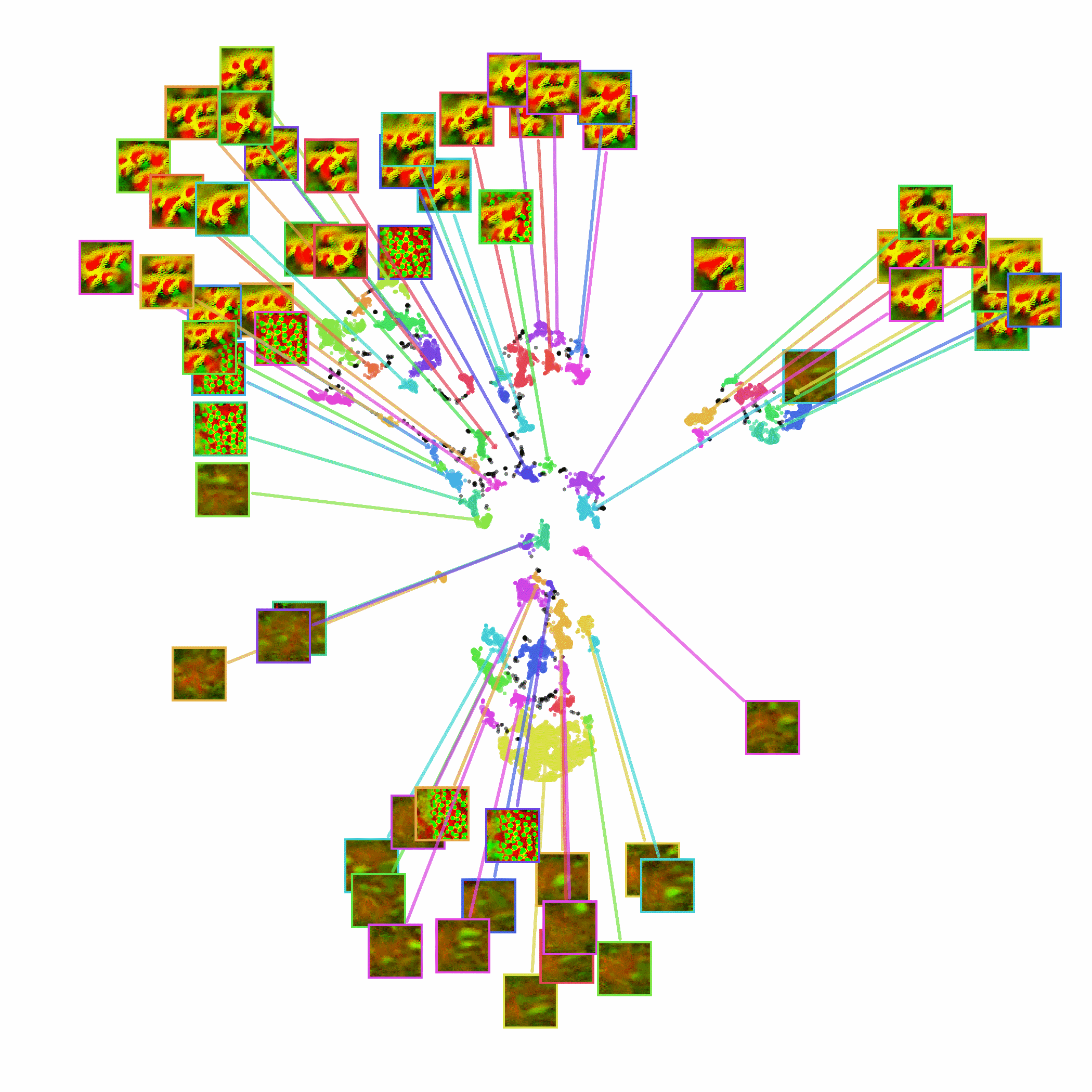} }}
\caption{\footnotesize Visualising clusters in two of the flattened feature tensor embeddings shown in figure \protect\ref{fig:tsne_flatten}. The clustering algorithm used was DBSCAN and the clusters are coloured in arbitrary hues. Points which were identified by the algorithm as not belonging to a cluster are coloured black. The feature images were generated using approach described in section \protect\ref{ssec:Visualising Clusters}.}
\label{fig:tsne_clusters_feature3}%
\end{figure}

To visualise a cluster we can generate a feature image based on the defining characteristics of the embedded feature tensors within the cluster. The method developed to do this compares the activation distribution of feature tensors within the cluster with feature tensors in the whole dataset. The difference between these two distributions was used to calculate a vector of channel weightings, $w_i$, of dimension $n_c$, where $n_c$ is the number of channels in the given layer. These weightings were then used the determine the degree to which each channel $i$ should be optimised for when generating a cluster's feature image. The pre-normalisation weights were calculated using the formula $ w_i =  \text{Relu}\big( ( m_{c,i} - m_{t,i} ) / \sigma_{t,i} \big) $, where $ m_{c,i}$ is the median channel-activation of channel $i$ for a cluster $c$, $ m_{t,i} $ is the median channel-activation of channel $i$ for total dataset, and $ \sigma_{t,i}  $ is the standard deviation of channel-activations of channel $i$ for total dataset. The weight vector was then normalised to be of unit length. Using this approach, two of the embeddings from figure \ref{fig:tsne_flatten} were visualised and are shown in figure \ref{fig:tsne_clusters_feature3}.

\section{Attribution \& Saliency Techniques}\label{S:4}

Attribution techniques look to understand how a network assembles the features it extracts and how these extracted features lead to classification decisions. One of the most common visualisations for attribution is called a \textit{saliency map} \cite{saliency}. This is a heatmap which highlights pixels of the input image that contribute most strongly to the output classification. Although we explored a number of attributed methods, in our limited usage we found it difficult to extract salient information with them, and as such no results using attribution methods are presented here. One reason for this may be due to the abstract nature of the class differences in our network which makes it difficult to determine whether or not the attribution information we obtain makes sense. Another reason could relate to the recent work done by Kindermans et$.$ al \cite{saliency-unreliable} which suggests that contemporary saliency methods and attribution techniques are not fully reliable. None the less, attribution methods do present a significant area for further work and are certainly something we would like to explore in the future.

\section{Conclusions}
\label{S:5}

The goal of this paper was to collate and develop the contemporary techniques for visualising the feature space of deep learning image classification neural networks. Of primary interest to us were dimension reduction visualisations and feature visualisation by optimisation techniques. Both of these areas were explored in detail and used to explicate the way in which our trained neural network distinguishes between cell types. To do this some novel techniques were developed, such as activation filtering, which can be used to extract the most salient information from single-channel feature images. Additionally, we tracked class separation in the network by embedding the flattened feature tensors from each layer of the network and applied clustering algorithms to these embedded spaces to elucidate structures within them using feature images. 

We believe much of this work and the visualisation techniques developed here could act as excellent supplements to contemporary scientific work, increasing the density of information presented while improving readability, as well as aid in the understanding of what is being learned by a convolutional neural network.

\section*{Acknowledgements}

This work was funded by the ImPACT program of the CSTI, Cabinet Office, Government of Japan and St. Catharine's College, Cambridge.

\end{document}